\documentclass[12pt]{amsart}
\usepackage{amsmath}
\usepackage{amsfonts,amssymb,amsthm}
\usepackage{graphicx}

\newcommand{\beq}{\begin{equation}}
\newcommand{\eeq}{\end{equation}}
\newcommand{\bbar}{\begin{eqnarray}}
\newcommand{\eear}{\end{eqnarray}}

\newcommand{\thm}[2]{\begin{#1} #2 \end{#1}}

\newtheorem{theorem}{Theorem}[section]
\newtheorem{itheorem}{Theorem}[section]
\newtheorem{lemma}[theorem]{Lemma}

\newtheorem{corollary}[theorem]{Corollary}

\newtheorem{remark}[theorem]{Remark}

\newtheorem{example}[theorem]{Example}
\newtheorem{definition}[theorem]{Definition}

\begin{document}

\title{The performance of the batch learning algorithm}

\author{Igor Rivin}
\address{Mathematics department, University of Manchester,
Oxford Road, Manchester M13 9PL, UK}
\address{Mathematics Department, Temple University,
Philadelphia, PA 19122}
\address{Mathematics Department, Princeton University, Princeton,
NJ 08544}
\email{irivin@math.princeton.edu} \thanks{The author would like 
to think the EPSRC and the NSF for support, and Natalia Komarova 
and  Ilan Vardi for useful conversations. }

\begin{abstract}
We analyze completely the convergence speed of the \emph{batch 
learning algorithm}, and compare its speed to that of the 
memoryless learning algorithm and of learning with memory (as 
analyzed in \cite{kr2}). We show that the batch learning 
algorithm is never worse than the memoryless learning algorithm 
(at least asymptotically). Its performance \emph{vis-a-vis} 
learning with full memory is less clearcut, and depends on 
certain probabilistic assumptions. 
\end{abstract}
\maketitle

\renewcommand{\theitheorem}{\Alph{itheorem}}
\section*{Introduction}
The original motivation for the work in this paper was provided 
by  research in learning theory, specifically in various models 
of language acquisition (see, for example, \cite{knn,nkn,kn}). In 
the paper \cite{kr2}, we had studied the speed of convergence of 
the  \emph{memoryless learner algorithm}, and also of 
\emph{learning with full memory}. Since the \emph{batch learning 
algorithm} is both widely known, and believed to have superior 
speed (at the cost of memory) to both of the above methods by 
learning theorists, it seemed natural to analyze its behavior 
under the same set of assumptions, in order to bring the analysis 
in \cite{kr1} and \cite{kr2} to a sort of closure. It should be 
noted that the detailed analysis of the batch learning algorithm 
is performed under the assumption of \emph{independence}, which 
was not explicitly present in our previous work. For the 
impatient reader we state our main result (Theorem 
\ref{batchthm}) immediately (the reader can compare it with the 
results on the memoryless learning algorithm and learning with 
full memory, as summarized in Theorem \ref{mainprev}): 
\begin{itheorem}
Let $N_\Delta$ be the number of steps it takes for the student 
to have probability $1 - \Delta$ of learning the
concept using the batch learner algorithm. Then we have the following estimates for $N_\Delta$:
\begin{itemize}
\item
if the distribution of overlaps is \emph{uniform}, or more 
generally, the density function $f(1-x)$  at $0$ has the form 
$f(x) = c + O(x^\delta),$ $\delta, c > 0,$ then there exist positive
constants $C_1, C_2$ such that 
$$\lim_{n \rightarrow \infty} \mathbf{P}\left(C_1 <
\frac{N_\Delta}{(1- \Delta)^2 n} < C_2\right) = 1$$
\item 
if the probability density function $f(1-x)$ is asymptotic to $c x^\beta
+ O(x^{\beta + \delta}), \quad \delta, \beta > 0$, as $x$ approaches
$0$, then 
$$\lim_{n \rightarrow \infty} \mathbf{P}\left(c_1 <
\frac{N_\Delta}{|\log \Delta|n^{\frac{1}{1+\beta}}} < c_2\right) = 1,$$
for some positive constants $c_1, c_2$;
\item 
if the asymptotic behavior is as above, but $-1 < \beta < 0$, then
$$\lim_{x \rightarrow \infty}  \mathbf{P}\left(\frac{1}{x} < \frac{N_\Delta}{|\log
\Delta| n^{1/(1+\beta)}} < x\right) = 1$$
\end{itemize}
\end{itheorem}
The plan of the paper is as follows: in this Introduction we 
recall the learning algorithms we study; in Section \ref{mathmod} 
we define our mathematical model; in Section 2 we recall our 
previous results, in Section 3 we begin the analysis of the batch 
learning algorithm, and introduce some of the necessary 
mathematical concepts; in Sections 4-6 we analyze the three cases 
stated in Theorem A, and we summarize our findings in Section 7.
\subsection*{Memoryless Learning and Learning with Full Memory} 
The general setup is as follows: There is a collection of 
concepts $R_0, \dots, R_n$ and words which refer to these 
concepts, sometimes ambiguously. The teacher generates a stream 
of words, referring to the concept $R_0$. This is not known to 
the student, but he must learn by, at each step, guessing some 
concept $R_i$ and checking for consistency with the teacher's 
input.  The \emph{memoryless learner algorithm} consists of 
picking a concept $R_i$ at random, and sticking by this choice, 
until it is proven wrong.  At this point another concept is 
picked randomly, and the procedure repeats. \emph{Learning with 
full memory} follows the same general process with the important 
difference that once a concept is rejected, the student never 
goes back to it. It is clear (for both algorithms) that once the 
student hits on the right answer $R_0$, this will be his final 
answer. We would like to estimate the probability of having 
guessed the right answer is after $k$ steps, and also the 
expected number of steps before the student settles on the right 
answer.

\subsection*{Batch Learning} The batch learning situation is 
similar to the above, but here the student records the words 
$w_1, \dots, w_k, \dots$ he gets from the teacher. For each word 
$w_i$ , we assume that the student can find (in his textbook, for 
example) a list $L_i$ of concepts referred to by the word. If we 
define 
\begin{equation*} 
\mathcal{L}_k = \bigcap_{i=1}^k L_i,
\end{equation*}
then we are interested in the smallest value of $k$ such that 
$\mathcal{L}_k = \{R_0\}$. This value $k_0$ is the time it has 
taken the student to learn the concept $R_0$. We think of $k_0$ 
as a random variable, and we wish to estimate its expectation.
\section{The mathematical model}
\label{mathmod}
 We think of the words referring to the concept 
$R_0$ as a probability space $\mathcal{P}$. The probability that 
one of these words also refer to the concept $R_i$ shall be 
denoted by $p_i$; the probability that a word refers to concepts 
$R_{i_1}, \dots, R_{i_k}$ shall be denoted by $p_{i_1 \dots 
i_k}$. All the results described below (obviously) depend in a 
crucial way on the $p_1, \dots, p_n$ and (in the case of the 
batch learning algorithm) also on the joint probabilities. Since 
there is no \emph{a priori} reason to assume specific values for 
the probabilities, we shall assume that all of the $p_i$ are 
themselves \emph{independent, identically distributed random 
variables}. We shall refer to their common distribution as 
$\mathcal{F}$, and to the density as $f$. It turns out that the 
convergence properties of the various learning algorithms depend 
on the local analytic properties of the distribution 
$\mathcal{F}$ at $1$ -- some moments reflection will convince the 
reader that this is not really so surprising. 

Sharper analysis of the batch learning algorithm, 
depends on the \emph{independence hypothesis}:
$$
p_{i_1 \dots i_k} = p_{i_1} \dots p_{i_k}.
$$
It is again not too surprising that some such assumption on 
correlations ought to be required for precise asymptotic results, 
though it is obviously the subject of a (non-mathematical) debate 
as to whether assuming that the various concepts are truly 
independent is reasonable from a cognitive science point of view. 

\section{Previous results}
In previous work \cite{kr1} and \cite{kr2} we obtained the 
following result. 
\thm{theorem}
{
\label{mainprev}
Let $N_\Delta$ be the number of steps it takes for the student 
to have probability $1 - \Delta$ of learning the
concept. Then we have the following estimates for $N_\Delta$:
\begin{itemize}
\item
if the distribution of overlaps is \emph{uniform}, or more 
generally, the density function $f(1-x)$  at $0$ has the form 
$f(x) = c + O(x^\delta),$ $\delta, c > 0,$ then there exist positive
constants $C_1, C_2, C_1', C_2'$ such that 
$$\lim_{n \rightarrow \infty} \mathbf{P}\left(C_1 <
\frac{N_\Delta}{|\log \Delta|n \log n} < C_2\right) = 1$$
for
the memoryless algorithm and
$$\lim_{n \rightarrow \infty} \mathbf{P}\left(C'_1 <
\frac{N_\Delta}{(1- \Delta)^2 n \log n} < C'_2\right) = 1$$
when learning with full memory;
\item 
if the probability density function $f(1-x)$ is asymptotic to $c x^\beta
+ O(x^{\beta + \delta}), \quad \delta, \beta > 0$, as $x$ approaches
$0$, then for the two algorithms we have respectively 
$$\lim_{n \rightarrow \infty} \mathbf{P}\left(c_1 <
\frac{N_\Delta}{|\log \Delta|n} < c_2\right) = 1,$$
and
$$\lim_{n \rightarrow \infty} \mathbf{P}\left(c_1' <
\frac{N_\Delta}{(1- \Delta)^2 n } < c_2'\right) = 1$$
for some positive constants $c_1, c_2, c_1', c_2'$;
\item 
if the asymptotic behavior is as above, but $-1 < \beta < 0$, then
$$\lim_{x \rightarrow \infty}  \mathbf{P}\left(\frac{1}{x} < \frac{N_\Delta}{|\log
\Delta| n^{1/(1+\beta)}} < x\right) = 1$$
for the memoryless learning algorithm, and similarly
$$\lim_{x \rightarrow \infty}  \mathbf{P}\left(\frac{1}{x} < \frac{N_\Delta}{(1-\Delta)^2 n^{1/(1+\beta)}} < x\right) = 1$$
for learning with full memory.
\end{itemize}}
\noindent Recall that $f(x) = \Theta(g(x))$ means that for 
sufficiently large $x$, the ratio $f(x)/g(x)$ is bounded between 
two strictly positive constants. The distribution of overlaps 
referred to above is simply the distribution $\mathcal{F}$. 
Notice that the theorem says nothing about the situation when 
$\mathcal{F}$ is supported in some interval $[0, a]$, for $a<1$. 
That case is (presumably) of scientific interest, but 
mathematically it is relatively trivial: we replace the arguments 
of all the $\Theta$s above by $1$, though, of course, we are 
thereby hiding the dependence on $a$.

\section{General bounds on the batch learner algorithm}

Consider a set of words $w_1, \dots, w_k$. The probability that 
they all refer to the concept $R_i$ is, obviously $p_i^k$. 
\begin{lemma}
\label{bounds}
 The probability $q_k$ that we still have not 
learned the concept $R_0$ after $k$ steps is bounded above by 
$\sum_{i=1}^n p_i^k$, and below by $\max_i p_i^k$. 
\end{lemma} 
\begin{proof}
Immediate. 
\end{proof}
We will first use these upper and lower 
bounds to get corresponding bounds on the convergence speed of 
the batch learner algorithm, and then invoke the independence 
hypothesis to sharpen these bounds in many cases.

We begin with a trivial but useful lemma.
\begin{lemma}
\label{rearrange}
 Let $G$ be a game where the probability of 
success (respectively failure) after at most $k$ steps is $s_k$ 
(respectively $f_k = 1-s_k $). Then the expected number of steps 
until success is 
$$\sum_{k=1}^\infty k (s_k - s_{k-1}) = \sum_{k=1}^\infty s_k = 1 - 
\sum_{k=1}^\infty f_k,$$ if the corresponding sum converges.
\end{lemma}
\begin{proof}
The proof is immediate from the definition of expectation and the 
possibility of rearrangment of terms of positive series.
\end{proof}
We can combine Lemma \ref{rearrange} and Lemma \ref{bounds} to 
obtain:
\begin{theorem}
\label{sumbounds} The expected time $T$ of convergence of the 
batch learner algorithm is bounded as follows:
\begin{equation}
\label{trivest} \sum_{i=1}^n \frac{1}{1-p_i} \geq T \geq 
\max_{1\leq i \leq n} \frac{1}{1-p_i}.
\end{equation}
\end{theorem}
The leftmost term in equation (\ref{trivest}) has been studied at 
length in \cite{kr1}. We state a version of the results of 
\cite{kr1} below:
\begin{theorem}
\label{allstab} Let $S=\sum_{i=1}^n \frac{1}{1-p_i},$ where the 
$p_i$ are independently identically distributed random variables 
with values in $[0, 1]$, with probability density $f$, such that 
$f(1-x) = x^\beta + O(x^{\beta + \delta}),\quad \delta > 0$ for 
$x\rightarrow 0$. Then If $\beta > 0$, then there exists a mean 
$m$, such that $\lim_{n \rightarrow \infty} \mathbb{P}(|S/n - m| 
> \epsilon) = 0,$ for any $\epsilon > 0.$ If $\beta = 0$, then 
$\lim_{n \rightarrow \infty} \mathbb{P}(|S/(n\log n) - 1| 
> \epsilon) = 0).$ Finally, if 
$-1 \leq \beta < 0,$ then $\lim_{n \rightarrow \infty} 
\mathbb{P}(S/n^{1/{\beta+1}} - C
> a) = g(a),$ where $\lim_{a \rightarrow \infty} g(a)= 0,$ and $C$ is 
an arbitrary (but fixed) constant, and likewise 
$$\mathbb{P}(S/n^{1/(\beta + 1)} < b) = h(b),$$ where $\lim_{a \rightarrow 0}h(a) = 0,$
\end{theorem}
The right hand side of Eq. (\ref{trivest}) is easier to 
understand. Indeed, let $p_1, \dots, p_n$ be distributed as usual 
(and as in the statement of Theorem \ref{allstab}). Then 
\begin{theorem}\label{expmin}
$$\lim_{n\rightarrow \infty} 
n^{\frac{1}{1+\beta}} \mathbf{E}\left(1-\max_{1 \leq i \leq n}
p_i\right) = C,$$
for some positive constant $C$.
\end{theorem}
\begin{proof}
First, we change variables to $q_i = 1 - p_i$. Obviously, the 
statement of the Theorem is equivalent to the statement that 
$$E = 
\mathbf{E}(\min_{1 \leq i \leq n} q_i) = C  n^{-1/{1+\beta}}.$$ We 
also write $h(x) = f(1-x),$ and let $H$ be the distribution function
whose density is $h,$ so that $H(x) = 1 - F(1-x).$
Now, the probability of that all of the $q_i$ are 
greater than  $t$ equals $1-(1-H(t))^n,$ so that 
$$E = \int_0^1 t~d\left[1-(1-H(t))^n\right] = \int_0^1 (1-H(t))^n d t.$$
We change variables $t = u/n^{1/(1+\beta)}$, to obtain
\begin{equation}
\label{firstint} E = \frac{1}{n^{1+\beta}} 
\int_0^{n^{\frac{1}{{1+\beta}}}} \left(1-H\left(\frac{u}{n^{1/(1+\beta)}}\right)\right)^n du. 
\end{equation}
Let us write $E = E_1(n) + E_2(n),$ where
\begin{gather}
\label{secondint}
E_1(n) = \int_0^{n^{\frac{1}{3 (\beta + 1)}}}
\left[1-H\left(\frac{u}{n^{1/(1+\beta)}}\right)\right]^n du,\\
E_2(n) = \int_{n^{\frac{1}{3 (\beta + 1)}}}^{n^{\frac{1}{1 + \beta}}}
\left[1-H\left(\frac{u}{n^{1/(1+\beta)}}\right)\right]^n du,
\end{gather}
Recall that
\begin{equation}
\label{asest}
H(x) = c x^{\beta+1} + O(x^{\beta + \delta + 1}).
\end{equation}
Let 
\begin{equation}
\label{eeint}
\mathcal{I} = \int_0^\infty \exp\left(c x^{1+\beta}\right) d x.
\end{equation}
We now show:
\begin{equation}
\label{secondint1}
\lim_{n \rightarrow \infty} E_1(n) = \mathcal{I}.
\end{equation}
This is an immediate consequence of Lemma \ref{explem} and Eq. (\ref{asest}).
Also,
\begin{equation}
\label{secondint2}
\lim_{n \rightarrow \infty} E_2(n) = 0.
\end{equation}
Since $H$ is a monotonically increasing function, it is sufficient to
show that
$$\lim_{n\rightarrow \infty} n^{\frac{1}{1 + \beta}} 
\left[1-H\left(n^{\frac{2}{3 (1 + \beta)}}\right)\right]^n = 0.$$
This is immediate from Eq. (\ref{asest}) and Lemma \ref{explem}.
\end{proof}
\begin{remark}
The argument shows that $C = \mathcal{I},$ where $C$ is the constant
in the statement of lemma, and $\mathcal{I}$ is the integral
introduced in Eq. (\ref{eeint}).
\end{remark}
\begin{lemma}
\label{explem}
Let $f_n(x) = (1-x/n)^n,$ and let $0 \leq z < 1/2.$ 
$$f_n(x) = \exp(-x)\left[1-\frac{x^2}{2 n} + O\left(\frac{x^3}{n^2}\right)\right].$$
\end{lemma}
\begin{proof}
Note that 
$$\log f_n(x) = n \log(1-x/n) = -x - \sum_{k=2}^\infty \frac{x^k}{kn^{k-1}}.$$
The assertion of the lemma follows by exponentiating the two sides of
the above equation.  
\end{proof}
We need one final observation:
\begin{theorem}
The variable $n^{1/(1+\beta)} \min_{i=1}^n q_i$ has a limiting 
distribution with distribution function $G(x) = 
1-\exp(-x^{1+\beta}).$
\end{theorem}
\begin{proof}
Immediate from the proof of Theorem \ref{expmin}.
\end{proof}

We can now put together all of the above results as follows.
\begin{theorem}
\label{allgen}
 Let $p_1, \dots, p_k$ be independently distributed 
with common density function $f$, such that $f(1-x) = c x^\beta + 
O(x^{\beta + \delta}),$ $\delta > 0$. Let $T$ be the expected 
time of the convergence of the batch learning algorithm with 
overlaps $p_1, \dots, p_k$. Then, if $\beta > 0$, then there 
exist $C_1, C_2$, such that  $C_1 n^{1/(1+\beta)} \leq T \leq C_2 
n$, with probability tending to $1$ as $n$ tends to $\infty$. If 
$\beta = 0$, then there exist $C_1, C_2$, such that $C_1 n \leq T 
\leq C_2 n \log n$, with probability tending to one as $n$ tends 
to $\infty.$ If $\beta > 0$, then $C^{-1} n^{1/(\beta + 1)} \leq 
T \leq C n^{1/(\beta + 1)}$ with probability tending to $0$ as 
$C$ goes to infinity.
\end{theorem}

The reader will remark that in the case that $\beta > 0$, the 
upper and lower bounds have the same order of magnitude as 
functions of $n$.

\section{Independent concepts}
We now invoke the independence hypothesis, whereby an application of the 
inclusion-exclusion principle gives us:

\thm{lemma}{\label{latmost} The probability $l_k$ that we have 
 learned the concept $R_0$ after $k$ steps is given by 
$$
l_k=\prod_{i=1}^n(1-p_i^k).
$$
}

Note that the probability $s_k$ of winning the game \emph{on the 
$k$-th step} is given by $s_k = l_k - l_{k-1}= (1-l_{k-1}) - 
(1-l_k)$. Since the expected number of steps $T$ to learn the 
concept is given by
$$T = \sum_{k=1}^\infty k s_k,$$
we immediately have  $$T = \sum_{k=1}^\infty (1-l_k)$$
\begin{lemma}
\label{letime} The expected time $T$ of learning the 
concept $R_0$ is given by
\begin{equation}
\label{letimeeq}
T = \sum_{k=1}^\infty \left(1-\prod_{i=1}^n 
\left(1-p_i^k\right)\right).
\end{equation}
\end{lemma}
Since the sum above is absolutely convergent, we can expand the 
products and interchange the order of summation to get the 
following formula for $T$:

\medskip\noindent
\textbf{Notation.}
Below, we identify subsets of $\{1, \dots, n\}$ with 
 multindexes (in the obvious way), and if $s = \{i_1, \dots, i_l\},$ then
$$p_s \stackrel{\mbox{def}}= p_{i_1} \cdots p_{i_l}.$$

\begin{lemma}
The expression Eq. (\ref{letimeeq}) can be rewritten as:
\begin{equation}
\label{subsum} T = \sum_{s\subseteq \{1, \dots, n\}} 
(-1)^{|s|-1} \left(\frac{1}{1-p_s} - 1\right),
\end{equation}
\end{lemma}

\begin{proof}
With notation as above,
\begin{equation*}
\prod_{i=1}^m \left(1-p_i^k\right) = 
\sum_{s \subseteq \{1, \dots, n\}} (-1)^{|s|} p_s^k,
\end{equation*}
so
\begin{equation*}
\begin{split}
T &= \sum_{k=1}^\infty \left(1 - \prod_{i=1}^n 
\left(1-p_i^k\right)\right)\\ 
&= \sum_{k=1}^\infty \left(1-\sum_{s \subseteq \{1, \dots, n\}}
(-1)^{|s|} p_s^k\right)\\
&= \sum_{s\subseteq \{1, \dots, n\}} (-1)^{|s|-1} 
\sum_{k=1}^\infty p_s^k \\
&= \sum_{s\subseteq \{1, \dots, n\}} 
(-1)^{|s|-1} \left(\frac{1}{1-p_s} - 1\right),
\end{split}
\end{equation*}
where the change in the order of summation is permissible since all
sums converge absolutely.
\end{proof}
Formula (\ref{subsum}) is useful in and of itself, but we now 
use it to analyse the statistical properties of the time of 
success $T$ under our distribution and independence assumptions. 
For this we shall need to study the \emph{moment zeta function} of a
probability distribution, introduced below. Its detailed properties
are investigated in my paper \cite{zeta}, where Theorems \ref{t1},
\ref{alpha1asymp} and \ref{alpha1asymp2}
below are proved. Below we summarize the definitions and the
results. 
\subsection{Moment zeta function}
\begin{definition}
\label{zdef} Let $\mathcal{F}$ be a probability 
distribution on a (possibly infinite) interval $I$, and let
$m_k(\mathcal{F}) =  \int_I x^k\mathcal{F}(d x)$ be the $k$-th moment
of  $\mathcal{F}$. Then the \emph{moment zeta function of  
$\mathcal{F}$} is defined to be $$\zeta_{\mathcal{F}}(s) =
\sum_{k=1}^\infty m_k^s(\mathcal{F}),$$ whenever the sum is defined. 
\end{definition}
The definition is, in a way, motivated by the following:

\begin{lemma}
\label{zetalemma} Let $\mathcal{F}$ be a probability 
distribution as above, and let $x_1, \dots, x_n$ be independent 
random variables with common distribution $\mathcal{F}$. Then
\begin{equation}
\mathbb{E}\left(\frac{1}{1-x_1 \dots x_n}\right) = 
\zeta_{\mathcal{F}}(n).
\end{equation}
In particular, the expectation is undefined whenever the zeta 
function is undefined. 
\end{lemma}
\begin{proof}
Expand the fraction in a geometric series and apply Fubini's 
theorem.
\end{proof}
\begin{example}
For $\mathcal{F}$ the uniform distribution on 
$[0, 1]$, $\zeta_{\mathcal{F}}$ is the familiar Riemann zeta 
function. 
\end{example}

Using standard techinques of asymptotic analysis, the following can be
shown (see \cite{zeta}):
\begin{theorem}
\label{momasymp}
Let $\mathcal{F}$ be a continuous distribution supported in $[0, 1],$ 
let $f$ be the density of the distribution $\mathcal{F}$, and
suppose that $f(1-x) = c x^\beta + O(x^{\beta + \delta}),$ for some
$\delta > 0.$ Then the $k$-th moment of $\mathcal{F}$ is asymptotic to 
$C k^{-(1+\beta)},$ for $C = c \Gamma(\beta).$ 
\end{theorem}

\begin{corollary}
Under the assumptions of Theorem \ref{momasymp}, 
$\zeta_{\mathcal{F}}(s)$ is defined for $s 
>1/(1+\beta)$.
\end{corollary}

The moment zeta function can be used to two of the three situations
occuring in the study of the batch learner algorithm:
In the sequel, we set $\alpha = \beta + 1$.
\subsection{$\alpha > 1$}
\label{isdef} 
In this case, we use our assumptions to rewrite Eq. 
(\ref{subsum}) as 
\begin{equation}
\label{subsum2} 
\mathbb{E}(T) = - \sum_{k=1}^n \binom{n}{k}(-1)^k \zeta_{\mathcal{F}}(k).
\end{equation}
This, in turn, can be rewritten (by expanding the definition of 
zeta) as
\begin{equation}
\label{subsum3} \mathbb{E}(T) = - \sum_{j=1}^\infty 
\left[\left(1-m_j(\mathcal{F})\right)^n-1\right] = 
\sum_{j=1}^\infty \left[1- \left(1-m_j(\mathcal{F})\right)^n\right]
\end{equation}

Using the moment zeta function we can show:
\begin{theorem}
\label{t1}
Let $\mathcal{F}$ be a continuous distribution supported on $[0, 1],$
and let $f$ be the density of $\mathcal{F}.$ Suppose further that 
$$\lim_{x \rightarrow 1} \frac{f(x)}{(1-x)^{\beta}} = c,$$ for $\beta,
c > 0.$ Then, 
\begin{equation*}
\begin{split}
\lim_{n\rightarrow \infty} n^{-\frac{1}{1+\beta}} \left[\sum_{k=1}^n
\binom{n}{k}(-1)^k \zeta_{\mathcal{F}}(k)\right] \\= 
-\int_0^\infty
\frac{1-\exp\left(-c\Gamma(\beta+1)u^{1+\beta}\right)}{u^2} du\\
= - \left(c \Gamma(\beta + 1)\right)^{\frac{1}{\beta+1}}
\Gamma\left(\frac{\beta}{\beta + 1}\right).
\end{split}
\end{equation*}
\end{theorem}
\subsection{$\alpha = 1$}
\label{medalpha} In this case, 
\begin{equation}
\label{asest02}
f(x) = L + o(1)
\end{equation} as $x$ 
approaches $1,$ and so Theorem \ref{momasymp} tells us that 
\begin{equation}
\label{asest2}
\lim_{j \rightarrow \infty} j m_j(\mathcal{F}) = L.
\end{equation}
It is not hard to see that 
$\zeta_{\mathcal{F}}(n)$ is defined for $n \geq 2$. We break up 
the expression in Eq. (\ref{subsum}) as 
\begin{equation}
\label{subsumm} T = \sum_{j=1}^n {\frac{1}{1-p_j} - 1} + 
\sum_{s\subseteq \{1, \dots, n\}, \quad |s| > 1} 
 (-1)^{|s|-1} 
\left(\frac{1}{1-p_s} - 1\right).
\end{equation}
Let 
\begin{gather*} T_1 = \sum_{j=1}^n {\frac{1}{1-p_j} - 1},\\
 T_2 = \sum_{s\subseteq \{1, \dots, n\}, \quad |s| > 1} 
 (-1)^{|s|-1} 
\left(\frac{1}{1-p_s} - 1\right).
\end{gather*}
 The first sum $T_1$ has 
no expectation, however $T_1/n$  does have have a stable 
distribution centered on $c \log n + c_2$. We will keep this in 
mind, but now let us look at the second sum  $T_2$. It can be 
rewritten as 
\begin{equation}
\label{subsumm2} T_2(n) = - \sum_{j=1}^\infty 
\left[\left(1-m_j(\mathcal{F})\right)^n-1 + n
m_j(\mathcal{F})\right]. 
\end{equation}
We can again use the moment zeta function to analyse the properties of
$T_2,$ to get:
\begin{theorem}
\label{alpha1asymp}
Let $\mathcal{F}$ be a continuous distribution supported on $[0, 1],$
and let $f$ be the density of $\mathcal{F}.$ Suppose further that 
$$\lim_{x \rightarrow 1} \frac{f(x)}{(1-x)} = c > 0.$$ 
Then, 
$$\sum_{k=2}^n
\binom{n}{k}(-1)^k \zeta_{\mathcal{F}}(k) \sim c n \log n.
$$
\end{theorem}
To get error estimates, we need stronger assumption on the function
$f$ than the weakest possible assumption made in Theorem
\ref{alpha1asymp}. 

\begin{theorem}
\label{alpha1asymp2}
Let $\mathcal{F}$ be a continuous distribution supported on $[0, 1],$
and let $f$ be the density of $\mathcal{F}.$ Suppose further that 
$$f(x) \sim c (1-x) + O\left((1-x)^\delta\right),$$ where $\delta > 0.$
Then, 
$$\sum_{k=2}^n
\binom{n}{k}(-1)^k \zeta_{\mathcal{F}}(k) \sim c n \log n + O(n).
$$
\end{theorem}

The conclusion differs somewhat from that of section
\ref{isdef} in that  we get an   
additional term of $c n \log n$, where $c = \lim_{x \rightarrow 
1} f(x) = \lim_{j \rightarrow \infty} j m_j$. This term is equal 
(with opposing sign) to the center of the stable law satisfied by 
$T_1$, so in case $\alpha = 1$, we see that $T$ has no 
expectation but satisfies a \emph{law of large numbers}, of the 
\begin{theorem}[Law of large numbers]
There exists a constant $C$ such that $\lim_{y \rightarrow 
\infty} \mathbf{P}(|T/n - C| > y) = 0.$
\end{theorem}
\section{$\alpha <1$}
\label{smallalpha} In this case the analysis goes through as in 
the preceding section when $\alpha > 1/2$, but then runs into 
considerable difficulties. However, in this case we note that 
Theorem \ref{allgen} actually gives us tight bounds. 
\section{The inevitable comparison}
We are now in a position to compare the performance of the batch 
learning algorithm with that of the memoryless learning algorithm 
and of learning with full memory, as summarized in Theorem 
\ref{mainprev}. We combine our computations above with the 
observation that the batch learner algorithm converges 
geometrically (Lemma \ref{latmost}), to get: 
\thm{theorem}
{
\label{batchthm}
Let $N_\Delta$ be the number of steps it takes for the student 
to have probability $1 - \Delta$ of learning the
concept using the batch learner algorithm. Then we have the following estimates for $N_\Delta$:
\begin{itemize}
\item
if the distribution of overlaps is \emph{uniform}, or more 
generally, the density function $f(1-x)$  at $0$ has the form 
$f(x) = c + O(x^\delta),$ $\delta, c > 0,$ then there exist positive
constants $C_1, C_2$ such that 
$$\lim_{n \rightarrow \infty} \mathbf{P}\left(C_1 <
\frac{N_\Delta}{(1- \Delta)^2 n} < C_2\right) = 1$$
\item 
if the probability density function $f(1-x)$ is asymptotic to $c x^\beta
+ O(x^{\beta + \delta}), \quad \delta, \beta > 0$, as $x$ approaches
$0$, then 
$$\lim_{n \rightarrow \infty} \mathbf{P}\left(c_1 <
\frac{N_\Delta}{|\log \Delta|n^{\frac{1}{1+\beta}}} < c_2\right) = 1,$$
for some positive constants $c_1, c_2$;
\item 
if the asymptotic behavior is as above, but $-1 < \beta < 0$, then
$$\lim_{x \rightarrow \infty}  \mathbf{P}\left(\frac{1}{x} < \frac{N_\Delta}{|\log
\Delta| n^{1/(1+\beta)}} < x\right) = 1$$
\end{itemize}}

%
Comparing Theorems \ref{mainprev} and \ref{batchthm}, we see that 
batch learning algorithm is uniformly superior for $\beta \geq 
0$, and the only one of the three to achieve \emph{sublinear} 
performance whenever $\beta 
> 0$ (the other two \emph{never} do better than linearly, unless 
the distribution $\mathcal{F}$ is supported away from $1.$) On 
the other hand, for $\beta < 0$, the batch learning algorithm 
performs comparably to the memoryless learner algorithm, and 
worse than learning with full memory.

\end{document}